%% file: document.tex
\documentclass[NewProceedings]{ascelike}

\usepackage[utf8]{inputenc}

\usepackage{amsmath, amssymb} 
\usepackage{bm} 
\usepackage{siunitx} 

\usepackage{hyperref} 

\usepackage{multirow} 

\usepackage{algorithm2e}
\SetKwInput{KwInput}{Input} 
\SetKwInput{KwOutput}{Output} 
\SetKwRepeat{Do}{do}{while}
\SetKwProg{throw}{throw}{\ Exception}{}

\usepackage{subfiles}

\usepackage{graphicx}
\graphicspath{{images/}} 
\usepackage{caption}
\usepackage{subcaption}
\usepackage{tikz}
\usetikzlibrary{fit}

\usepackage[colorinlistoftodos]{todonotes} 
\presetkeys{todonotes}{inline}{} 

\usepackage[style=authoryear,giveninits=true,maxcitenames=1]{biblatex}
\title{Soft Tensegrity Systems for Planetary Landing and Exploration}
\author{
K\'evin Garanger$^1$
\and
Isaac del Valle$^1$
\and 
Miriam Rath$^1$
\and 
Matthew Krajewski$^1$
\and 
Utkarsh Raheja$^1$
\and 
Marco Pavone$^2$
\and 
Julian J. Rimoli$^1$
\and
\\
\parbox[t]{5.75in}{
    $^1$School of Aerospace Engineering, Georgia Institute of Technology \\
    Atlanta, GA 30332 \\
    $^2$Department of Aeronautics and Astronautics, Stanford University \\
    Stanford, CA 94305
}
}

\begin{document}

\maketitle

\begin{abstract}
During the last decade, tensegrity systems have been the focus of numerous investigations exploring the possibility of adopting them for planetary landing and exploration applications. Early approaches mainly focused on locomotion aspects related to tensegrity systems, where mobility was achieved by actuating the cable members of the system. Later efforts focused on understanding energy storage mechanisms of tensegrity systems undergoing landing events. More precisely, it was shown that under highly dynamic events, buckling of individual members of a tensegrity structure does not necessarily imply structural failure, suggesting that efficient structural design of planetary landers could be achieved by allowing its compression members to buckle. In this work, we combine both aspects of previous research on tensegrity structures, showing a possible lattice-like structural configuration able to withstand impact events, store pre-impact kinetic energy, and utilize a part of that energy for the locomotion process. Our work shows the feasibility of this proposed approach via both experimental and computational means.
\end{abstract}


\section{Introduction}

\subfile{sections/introduction}

\section{Proposed architecture and modeling approach}\label{sec:architecture}

\subfile{sections/architecture}

\section{Simulation results}\label{sec:simulation}

\subfile{sections/simulation}

\section{Experiments results}\label{sec:experiments}

\subfile{sections/experiments}

\section{Conclusion}\label{sec:conclusion}
\subfile{sections/conclusion}

\section{Acknowledgments}
Marco Pavone and Julian J. Rimoli would like to acknowledge the support provided by the National Academy of Engineering through the Grainger Foundation Frontiers of Engineering Grant, which made this work possible.

K\'evin Garanger would like to acknowledge his financial support thoughout this work from King Abdullah University of Science and Technology through Visiting Student Research Program and the Air Force Research Laboratory Materials and Manufacturing Directorate.

\printbibliography

\end{document}

%% file: sections/introduction.tex
A tensegrity system is an actuated structure made of tension and compression elements arranged in such a way that tensile forces stabilize the structure. Given their light weight and ability to withstand high velocities impacts, tensegrity systems have been the subject of several studies considering the possibility of adopting them for planetary landing and exploration applications \autocite{sabelhaus2015system}. Most approaches focus on the locomotion of tensegrity systems after landing by actuating the cable members to achieve a rolling or gait-like motion \autocite{paul2006design, sabelhaus2015system, chen2017soft, vespignani2018design}. In a recent work,  \autocite{kim2016hopping} suggests combining this type of locomotion with a thruster to hop over long distances.

These works are part of an exciting and broader field of aerospace research concerned with the design of efficient rovers for the exploration of Solar System bodies. The low-gravity environment and rough surfaces characteristic of small Solar System bodies offer new possibilities and challenges that must be met with clever designs that rely on different modes of locomotion than the traditional wheeled rover, which is inappropriate for such environments \autocite{pavone2013spacecraft, hockman2017design}.

When it comes to modeling tensegrity systems, typical approaches do not account for the buckling of compression members. The loads borne by individual members are therefore limited to stay under their Euler's critical load so that the modeling approach remains valid. While this is a reasonable assumption when modeling the rolling locomotion of spherical tensegrity systems, it was shown to be inaccurate for highly dynamic events such as planetary landings \autocite{rimoli2016impact, rimoli2018reduced}. Allowing compression members to buckle could lead to more efficient structural designs able to store pre-impact kinetic energy and utilize part of it for the locomotion process. In our work, we consider a tensegrity system with a lattice-like configuration made of truncated octahedron elementary cells. This architecture opens the way to modular designs with the possibility to build complex lattice structures with various shapes \autocite{rimoli2017mechanical}. The type of locomotion presented in this paper is far different from the motions usually achieved with spherical tensegrity systems \autocite{chen2017soft, vespignani2018design}, as it is more akin to hopping than rolling. The study of this type of locomotion is made possible by the reduced-order model of the tensegrity structure that enables the analysis of the energy storage and release mechanisms through the buckling of the compression members.

In section \ref{sec:architecture}, a description of the tensegrity architecture used in this work is made and the reduced-order model used for simulating the tensegrity system is summarized. Section \ref{sec:simulation} presents the results obtained from the simulation and the approach used to learn a hopping mode of locomotion for the tensegrity system. The results from transferring these findings to a real prototype are given in section \ref{sec:experiments}. A conclusion is finally drawn from those results in section \ref{sec:conclusion}.

%% file: sections/architecture.tex
\subsection{Architecture}

\subsubsection{Unit cell}

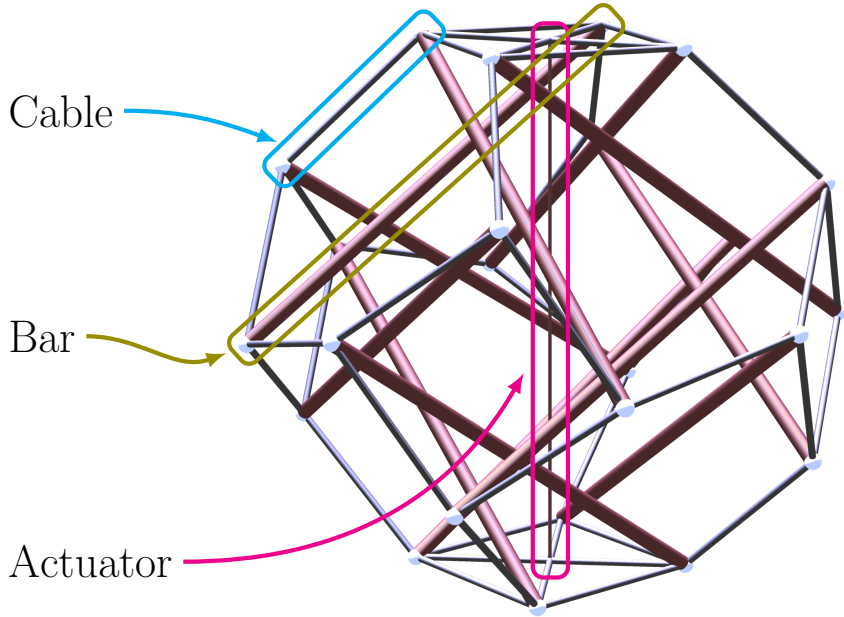
\begin{figure}[h]
\centering
\begin{tikzpicture}
\node [anchor=west] (bar) at (0,4) {\Large Bar};
\node [anchor=west] (cable) at (0,7) {\Large Cable};
\node [anchor=west] (actuator) at (0,1) {\Large Actuator};
\begin{scope}[xshift=1.5cm]
    \node[anchor=south west, inner sep=0] (image) at (0,0) {\includegraphics[width=0.8\textwidth]{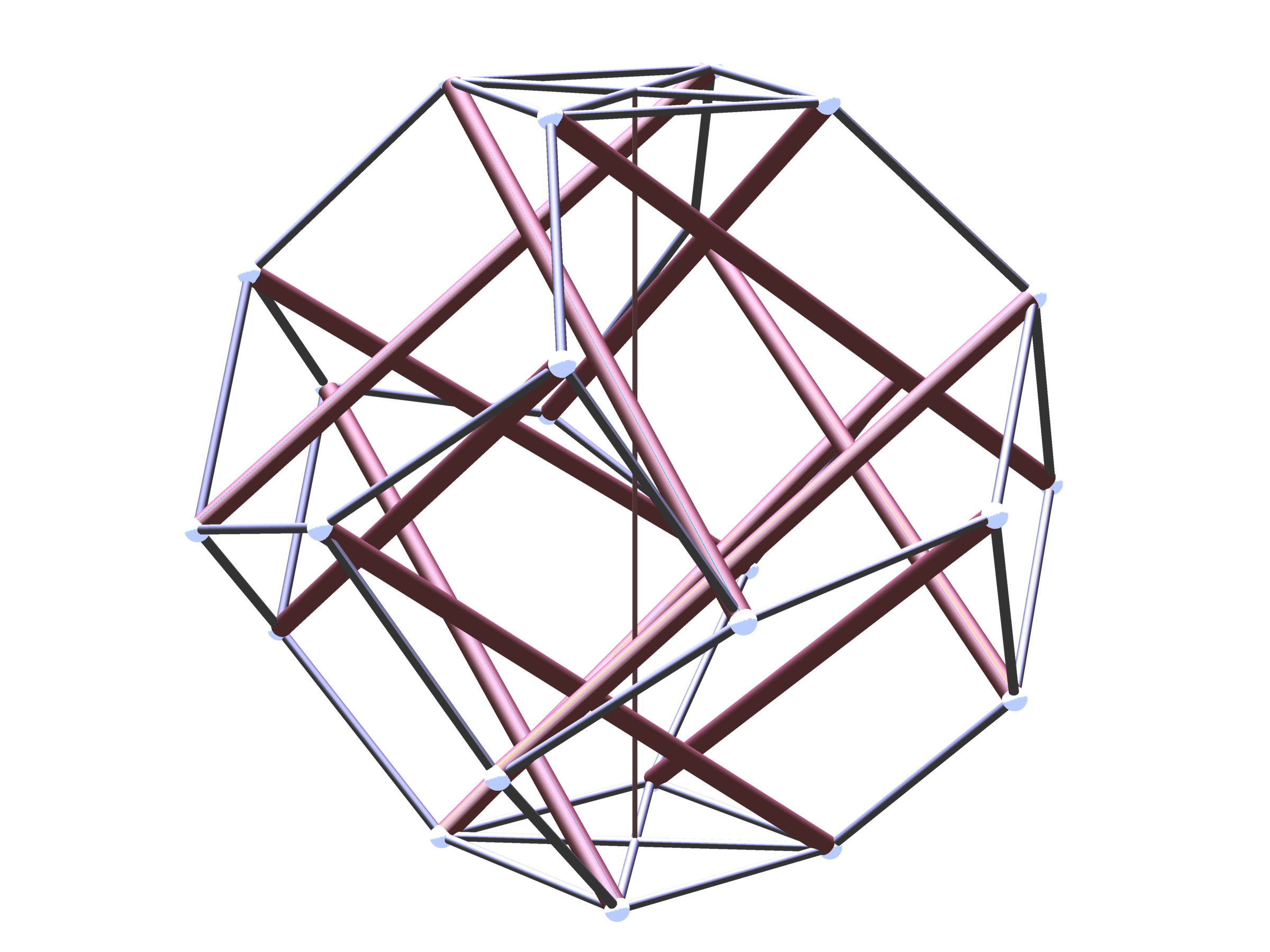}};
    \begin{scope}[x={(image.south east)},y={(image.north west)}]
        \draw[magenta,ultra thick,rounded corners] (0.48,0.09) rectangle (0.52,0.93);
        \draw [-latex, ultra thick, magenta] (actuator) to[out=0, in=-120] (0.47  	,0.40);
        \draw[cyan,ultra thick,rounded corners,rotate around={-46:(0.17,0.715)}] (0.17,0.715) rectangle (0.21,1.06);
        \draw [-latex, ultra thick, cyan] (cable) to[out=0, in=-200] (0.19,0.75);
        \draw[olive,ultra thick,rounded corners,rotate around={-48:(0.13,0.447)}] (0.13,0.447) rectangle (0.17, 1.22);
        \draw [-latex, ultra thick, olive] (bar) to[out=0, in=-160] (0.132,0.43);
    \end{scope}
\end{scope}
\end{tikzpicture}

\caption{Rendering of the truncated octahedron unit cell.}
\label{fig:unit_cell_plot}
\end{figure}

The architecture proposed in this work is based on a lattice made from the repetition of a truncated octahedron unit cell. Each unit cell is made of \si{12} bars and \si{44} unactuated cables. The bar ends, where the cables connect, are referred to as nodes, of which there are \si{24} total. \si{36} of the unactuated cables correspond to the edges of the truncated octahedron. Some of these cables are shared between two unit cells in a lattice configuration. The other \si{8} cables are added to attach both ends of an actuated cable linking the top and bottom faces of each unit cell. This actuated cable can be retracted around a reel that can be locked to prevent the cable from unwinding. As the actuated cable is retracted, the structure is compressed and accumulates elastic energy, mostly from the bending of the compression members. The reel can then be unlocked in order to quickly unwind the actuated cable and release the elastic energy stored in the structure, initiating a jump. See Fig.~\ref{fig:unit_cell_plot} for a graphical depiction of the elements of an elementary cell at rest.

\subsubsection{Lattice assembly}

Although able to hop in place, a single elementary cell cannot be steered toward a given arbitrary direction with the proposed actuation mechanism. A possible solution is to use multiple elementary cells attached together, to retract the actuated cables to different lengths, and to release them at the same time to create a jump in the desired direction.
The lattice assembly of the unit cells is performed as described in \autocite{rimoli2017mechanical}, where two adjacent unit cells are symmetrical about the plane at their intersection. Because of the symmetries of the unit cell, the reflection from one unit cell to an adjacent one is equivalent to the composition of a translation and a rotation. All unit cells of the lattice are therefore similar despite the reflection operation used to generate the lattice. This configuration differs from the one of traditional lattices, which are normally made by a pure translation of the unit cell. Indeed, the non symmetry of opposing faces in the unit cell makes the construction of a lattice by pure translations impossible. With the used lattice, bars connect at the intersection between unit cells and are grouped in connected components of at most four bars forming a loop. This architecture is therefore in accordance with the fundamental concept of tensegrity stating that compression elements form islands of compression in a sea of tension \autocite{buckminster1962tensile}.

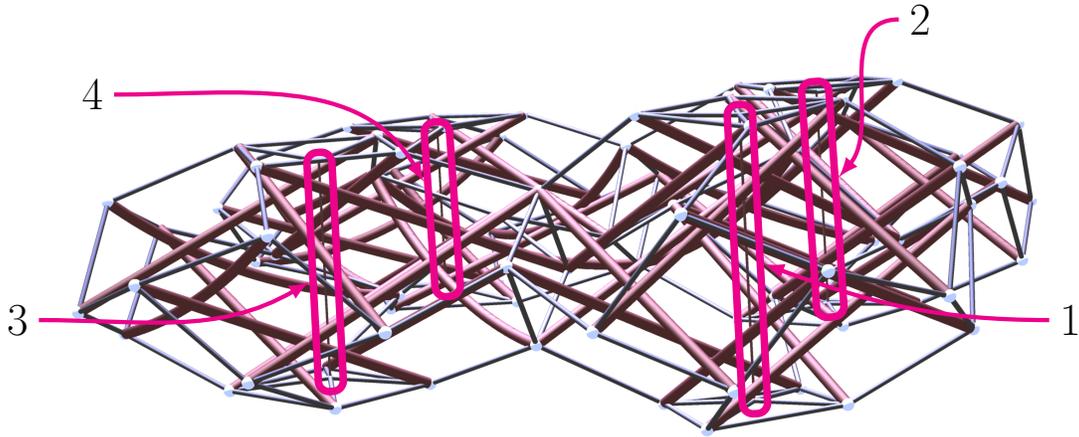
\begin{figure}[h]
\centering
\begin{tikzpicture}
\node [anchor=west] (act1) at (14, 4) {\Large $1$};
\node [anchor=west] (act2) at (12, 8) {\Large $2$};
\node [anchor=west] (act3) at (0,4) {\Large $3$};
\node [anchor=west] (act4) at (1,7) {\Large $4$};
\begin{scope}[xshift=0.5cm]
    \node[anchor=south west, inner sep=0] (image) at (0,0) {\includegraphics[width=\textwidth]{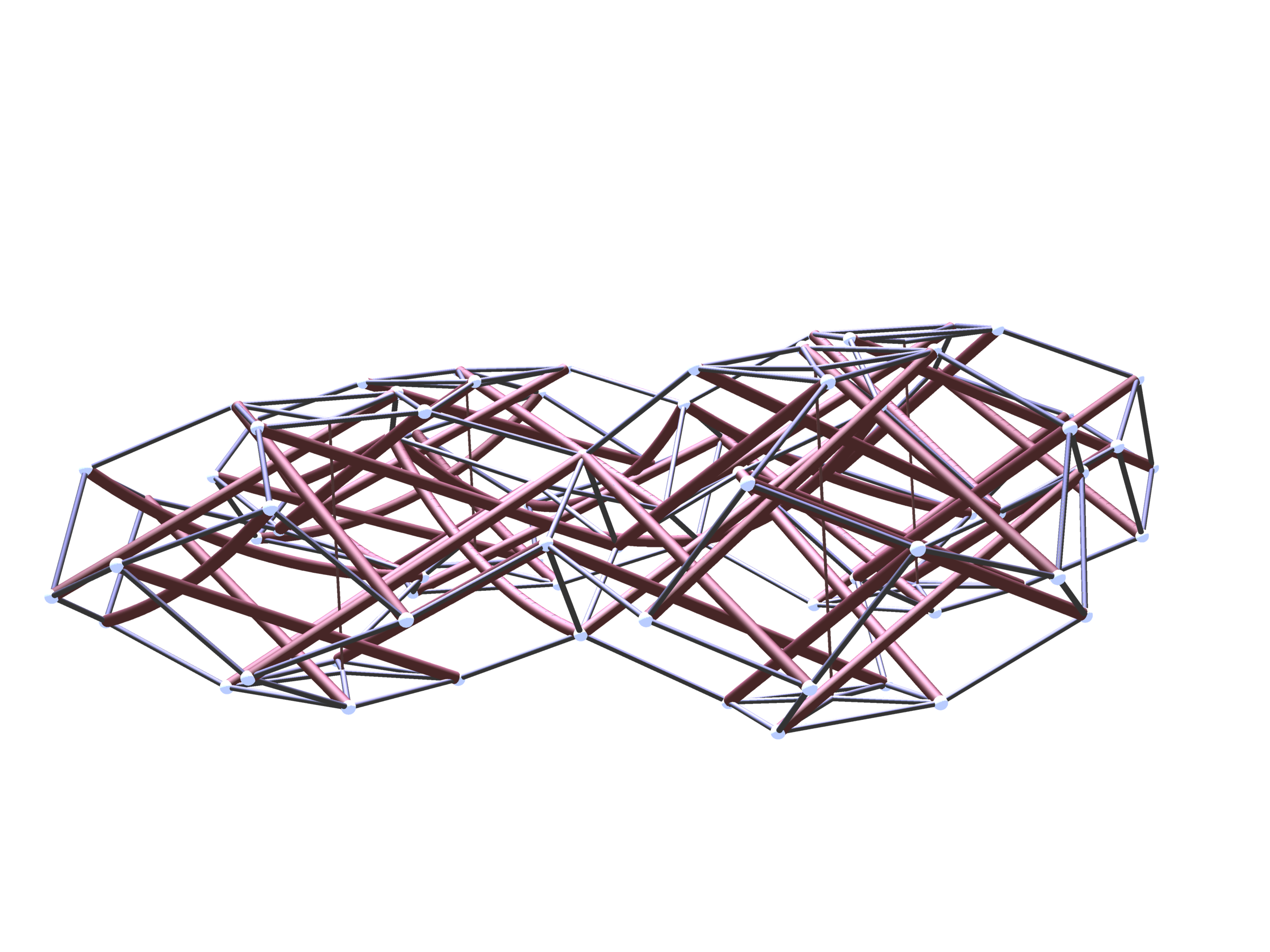}};
    \begin{scope}[x={(image.south east)},y={(image.north west)}]
		\node[anchor=south] (p1) at (0.27, 0.28) {};
		\node[anchor=south] (p2) at (0.26, 0.54) {};
		\node[anchor=south] (p3) at (0.376, 0.395) {};
		\node[anchor=south] (p4) at (0.367, 0.577) {};
		\node[anchor=south] (p5) at (0.656, 0.253) {};
		\node[anchor=south] (p6) at (0.642, 0.598) {};
		\node[anchor=south] (p7) at (0.723, 0.37) {};
		\node[anchor=south] (p8) at (0.712, 0.626) {};

        \node[draw=magenta,line width=1mm,rounded corners, inner sep=0.2mm, rotate fit=3, fit=(p5) (p6)] (a1) {};
        \node[draw=magenta,line width=1mm,rounded corners, inner sep=0.2mm, rotate fit=4, fit=(p7) (p8)] (a2) {};
        \node[draw=magenta,line width=1mm,rounded corners, inner sep=0.2mm, rotate fit=3, fit=(p1) (p2)] (a3) {};
        \node[draw=magenta,line width=1mm,rounded corners, inner sep=0.2mm, rotate fit=4, fit=(p3) (p4)] (a4) {};
        
        \draw [-latex, ultra thick, magenta] (act1) to [out=180, in=-20] (a1);
		\draw [-latex, ultra thick, magenta] (act2) to [out=180, in=50] (a2);
		\draw [-latex, ultra thick, magenta] (act3) to [out=0, in=-150] (a3);
        \draw [-latex, ultra thick, magenta] (act4) to [out=0, in=-240] (a4);
    \end{scope}
\end{scope}
\end{tikzpicture}
\caption{Indexing of the actuators for the $2\times 2$ lattice.}
\label{fig:2x2_plot}
\end{figure}

In this work we focus on the control of a $2\times 2$ structure since it is the smallest assembly capable of hopping in any direction given the actuation mechanism. This structure, shown in Fig.~\ref{fig:2x2_plot}, is made of \si{80} nodes, \si{48} bars, \si{160} cables, and \si{4} actuators.

\subsubsection{Prototype}

To assess the viability of the hopping locomotion in practice, a prototype of the $2\times 1$ tensegrity lattice was built and used for the experiments of section \ref{sec:experiments}. The bars of the prototype are made of carbon fiber and the cables of stainless steel.
At both ends of each bar, a ball joint connects the bar to a 3D printed part to which cables can attach.
The actuation mechanism is also attached to the 3D printed parts using the additional cables. The mechanism consists of a reel that is connected to a constant force spring. When the unit is compressed, the constant force spring retracts the actuated cable, driving the reel past a spring-loaded lock. The interaction between the reel and lock is similar to that of a ratchet. When engaged, the lock allows the reel to rotate in one direction only, preventing the unwinding of the actuated cable. Upon reaching the desired accumulated elastic energy, a servo disengages the lock from the reel, allowing the actuated cable to rapidly unwind and release the energy of the unit, initiating a jump.

\begin{figure}[h]
\centering
\includegraphics[width=0.7\textwidth]{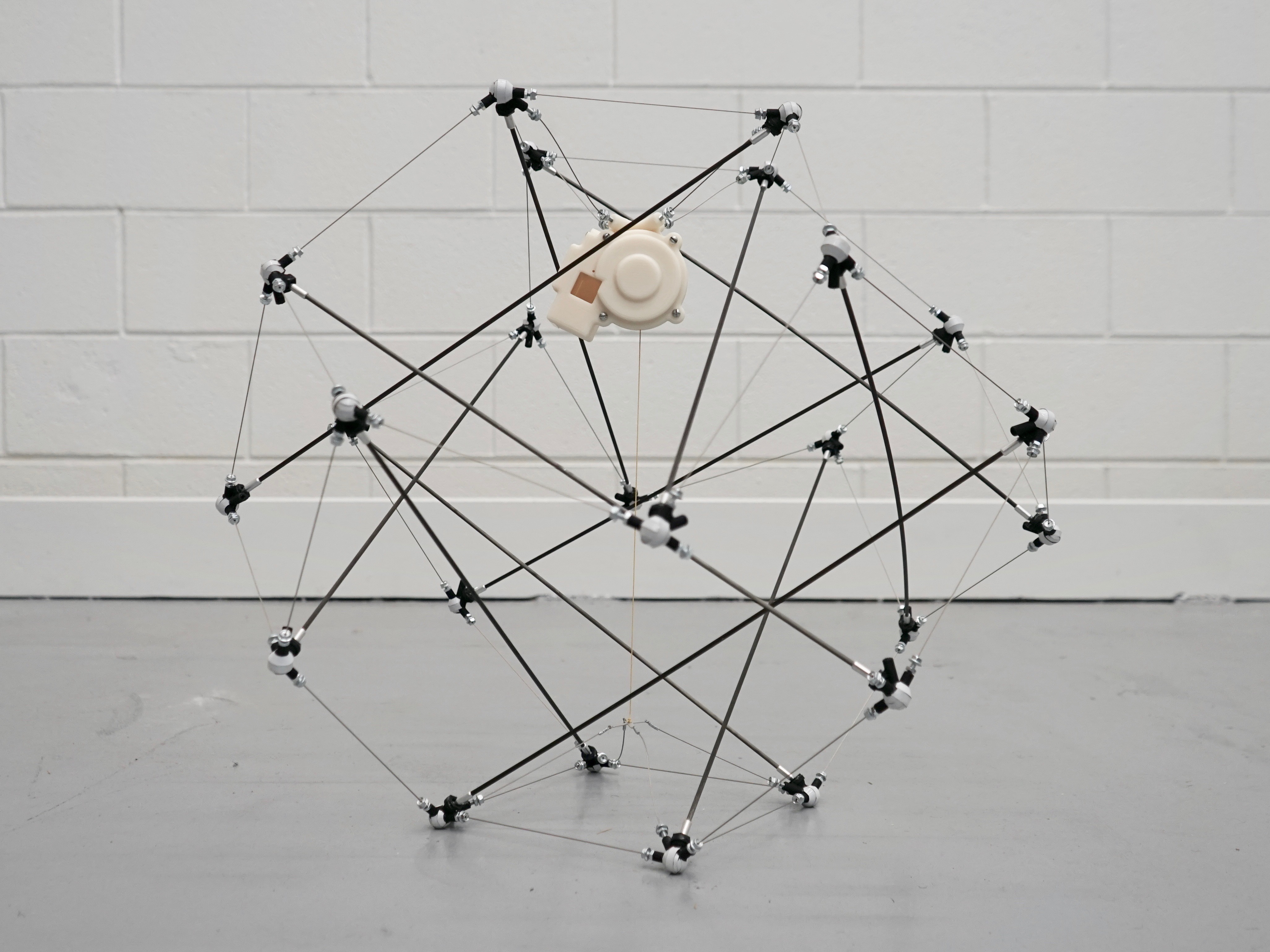}
\caption{Picture of the unit cell prototype.}
\label{fig:unit_cell_picture}
\end{figure}

\subsubsection{State estimation}

The state estimation of the prototype is performed with a Vicon camera three-dimensional motion tracking system using Nexus data capture software. Each node on the lattice is covered with reflective tape to represent a marker inside the software. Experiments, in which the lattice is compressed and jumps, are performed using \si{12} cameras with a 1000Hz collection rate. As the system tracks the trajectory of each individual node, it is possible to estimate the trajectory of the center of mass of the lattice by averaging the coordinates of the nodes at each time instance. This allows the center of mass to be tracked through three-dimensional space once a jump is completed.

\subsection{Modeling approach}

In order to use the hopping motion of the tensegrity lattice as a viable mode of locomotion, it seems reasonable to expect the ability to predict the landing point after a hop for a given actuator configuration and environment. The environment includes the factors that have a significant impact on the hopping trajectory such as gravity and terrain. In this study, we limit ourselves to a flat and horizontal terrain.

To estimate the hopping trajectory of the tensegrity system for a given actuator configuration, a simulation based on a reduced-order physical model is built. The simulation is based on the integration of the physical model dynamics using explicit methods. For the simulation to remain stable, the integration time step has to be of the same order of magnitude as the inverses of the natural frequencies of the spring elements. This time step gets smaller as the simulated structures grow stiffer which implies an increase in the computational time required for the simulation. For the structures considered in this work, the simulation runs slower than real-time by several orders of magnitude. Simulating the physical model for real-time motion planning is therefore not a possible approach. However, a dataset of simulated trajectories can be constituted and used to learn the hopping behavior of the tensegrity system. In Section \ref{sec:simulation}, the constitution of such a dataset is described.

\subsubsection{Physical model of the structure}

\begin{figure}[h]
\centering
\includegraphics[width=\textwidth]{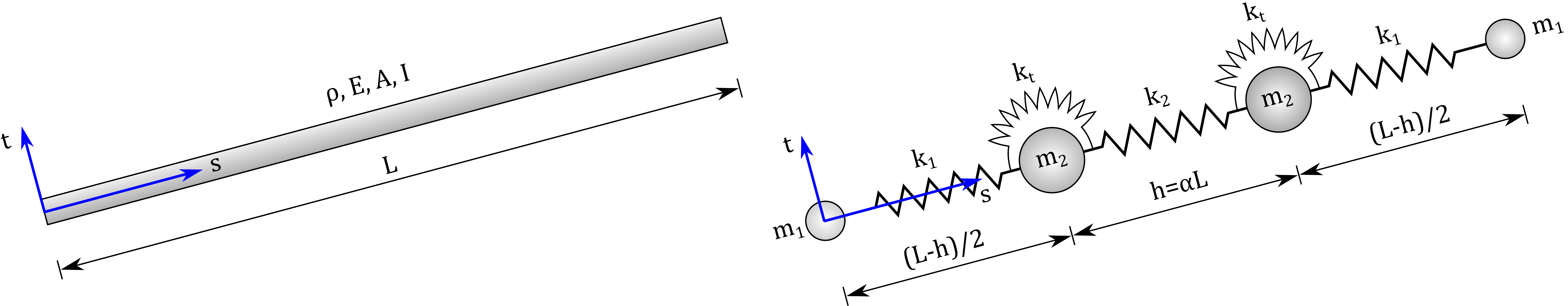}
\caption{Schematic of the discretization scheme of bars. The continuous bar on the left is replaced by the discretized system on the right.}
\label{fig:discretized_bar}
\end{figure}

The physical model used in this work for the simulation of the tensegrity system is based on the reduced-order model introduced in \autocite{rimoli2018reduced}. Cables and actuators are modeled by linear springs that only produce a force when in tension while bars are discretized in a set of four masses, three linear springs, and two angular springs as depicted in Fig.~\ref{fig:discretized_bar}. Two point masses are added at each end of each actuated cable, for a total of \si{184} point masses constituting the modeled state of the $2\times 2$ tensegrity lattice. The weight of each cable is shared between its two adjacent masses. This discretization scheme accurately approximates the buckling of the compression elements with only a reduced number of parameters.

\subsubsection{Contact forces}

Apart from gravity, the only other external forces considered in the model are the ones induced by the contact of the tensegrity system with the ground. They include the normal reaction force and friction force experienced by each node during contact or collision with the ground. The reaction and friction forces are modeled by using a collision impulse model.

\subsection{Simulation of the hopping behavior}

\subsubsection{Initial state form-finding}

The determination of the equilibrium of the studied tensegrity system for a given set of actuators rest lengths, often referred as form-finding, can be done by using different methods. The two methods that were considered in this work are dynamic relaxation and conjugate gradient descent. In dynamic relaxation the system is simulated from an initial state with the rest lengths adjusted to the desired value until an equilibrium is reached. A viscous damping term is added to each node to speed up the convergence. Although able to systematically find the equilibrium configuration position of a tensegrity system, the dynamic relaxation method is too slow to be practicable to generate a high number of equilibrium configurations. In the performed simulations, the conjugate gradient descent was much faster than dynamic relaxation, by roughly two orders of magnitude.

In conjugate gradient descent, an energy minimum is searched by iteratively updating the state while following a direction updated at each iteration with the energy gradient of the current state. Whereas in a steepest descent the direction is always equal to the opposite of the energy gradient, in a conjugate gradient descent the direction $c_n$ at step $n > 0$, called the conjugate, is computed according to the following formula:
\[
    c_n = -\nabla_{x} E(x_n) + \beta_n c_{n-1},
\]
where $x_n$ is the state at step $n$, $E(x_n)$ the energy of the state, and $\beta_n$ a real scalar. $c_0$ is set by following the opposite of the gradient direction at $x_0$. 

A steepest descent would be equivalent to setting $\beta_n = 0$ at each step. In our implementation, $\beta_n$ was computed according to the Polak-Ribi\'ere formula~\autocite{polak1969note} and clipped to always ensure directions implying sufficiently negative slopes.
Once the conjugate $c_n$ is computed, a line search is performed to find the step $\alpha_n$ that minimizes $E(x_n + \alpha_n c_n)$ and to update the state. The conjugate is then updated and the process repeated until convergence. Many different possibilities exist to implement the line search and the best performing ones usually depend heavily of the problem study. In general, there is a trade-off to consider between refining the line search to obtain a better minimum of the energy along the conjugate direction at a computational cost and performing a step with a suboptimal estimate of the energy minimum along the searched line.

In our case, since the variation of energy of a tensegrity lattice along a given direction is usually dominated by the elastic energy of the system, the quasi-quadratic nature of the energy is used to quickly find an energy minimum, as described in more details in the following.

When proceeding to the line search at state $x$ for a given direction $c$, we write $E_\alpha = E(x + \alpha c)$.
The line search proceeds by first finding three successive nonnegative scalars $\alpha_0$, $\alpha_1$, and $\alpha_2$ such that the energy at state $E_{\alpha_1} < E_{\alpha_0}$ and $E_{\alpha_1} < E_{\alpha_2}$, indicating the presence of a local minimum of the energy between $\alpha_0$ and $\alpha_2$. Then, a new $\alpha_m \in (\alpha_0, \alpha_2)$ is computed as the absissa where the parabola intercepting the points $(\alpha_0, E_{\alpha_0})$, $(\alpha_1, E_{\alpha_1})$, and $(\alpha_2, E_{\alpha_2})$ reaches its minimum. The following update rule is then applied:
\begin{equation*}
    (\alpha_0, \alpha_1, \alpha_2) = \left\{
  \begin{array}{@{}ll@{}}
      (\alpha_0, \alpha_m, \alpha_1), & \text{if}\ \alpha_m < \alpha_1\ \text{and}\ E_{\alpha_m} < E_{\alpha_1} \\
      (\alpha_m, \alpha_1, \alpha_2), & \text{if}\ \alpha_m < \alpha_1\ \text{and}\ E_{\alpha_m} > E_{\alpha_1} \\
      (\alpha_1, \alpha_m, \alpha_2), & \text{if}\ \alpha_1 < \alpha_m\ \text{and}\ E_{\alpha_m} < E_{\alpha_1} \\
      (\alpha_0, \alpha_1, \alpha_m), & \text{if}\ \alpha_1 < \alpha_m\ \text{and}\ E_{\alpha_m} > E_{\alpha_1}
  \end{array}\right.
\end{equation*} 
Is is easy to show that the new interval $[\alpha_0, \alpha_1]$ still contains a local minimum of the energy. This process is then repeated until a minimum is found with an acceptable tolerance or when the number of iterations reaches a given maximum. $\alpha_1$ is then returned by the line search and the state is updated accordingly.

%% file: sections/simulation.tex
\subsection{Simulation analysis}

Simulations of the tensegrity system behavior were performed for each of the equilibrium configuration found previously via conjugate gradient descent.
More than \si{40000} configurations were computed for randomized actuators rest lengths. The rest lengths were sampled between $0.2l$ and $0.8l$, where $l$ is the distance between two opposed face of the truncated octahedron formed by a unit cell subject to no external load. That is, a rest length $0.2l$ corresponds to a stretch $\lambda=0.2$, since $\lambda$ is defined as the ratio between the rest and original length of the cable.

\begin{figure}[!ht]
	\centering
	\begin{tikzpicture}
	\node[anchor=south west, inner sep=0] (structure) at (0,0) {\includegraphics[width=\textwidth]{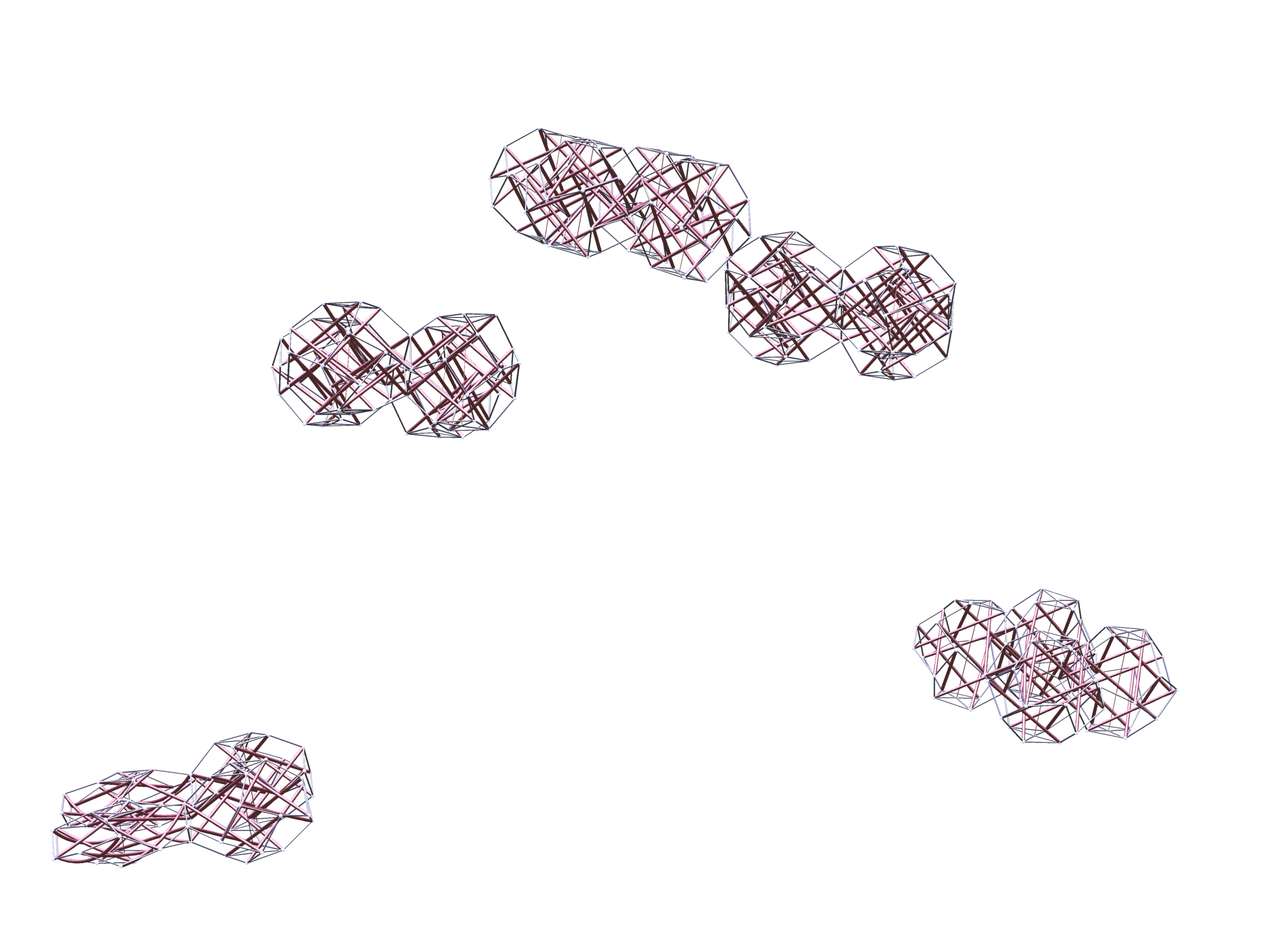}};
	\node[anchor=south west, inner sep=0] (com) at (0,0) {\includegraphics[width=\textwidth]{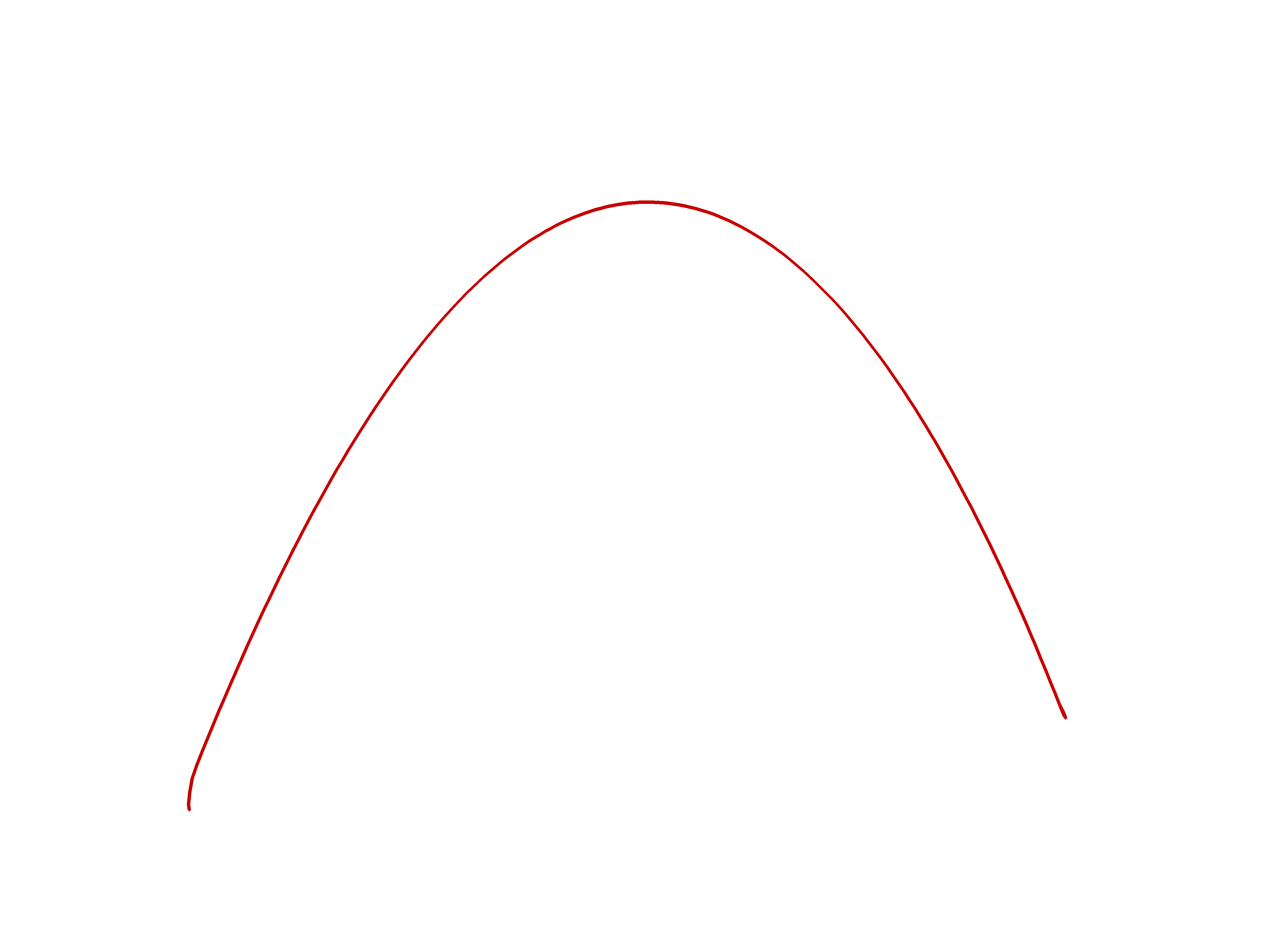}};
	\begin{scope}[x={(image.south east)},y={(image.north west)}]
	\node[anchor=south] (p0) at (0.05, 0.2) {\SI{0.0}{\second}};
	\node[anchor=south] (p0) at (0.25, 0.7) {\SI{0.4}{\second}};
	\node[anchor=south] (p0) at (0.45, 0.88) {\SI{0.8}{\second}};
	\node[anchor=south] (p0) at (0.7, 0.75) {\SI{1.2}{\second}};
	\node[anchor=south] (p0) at (0.82, 0.4) {\SI{1.6}{\second}};
	\end{scope}
	\end{tikzpicture}
	\caption{Rendering of the hopping trajectory of the tensegrity system.}
	\label{fig:jump}
\end{figure}
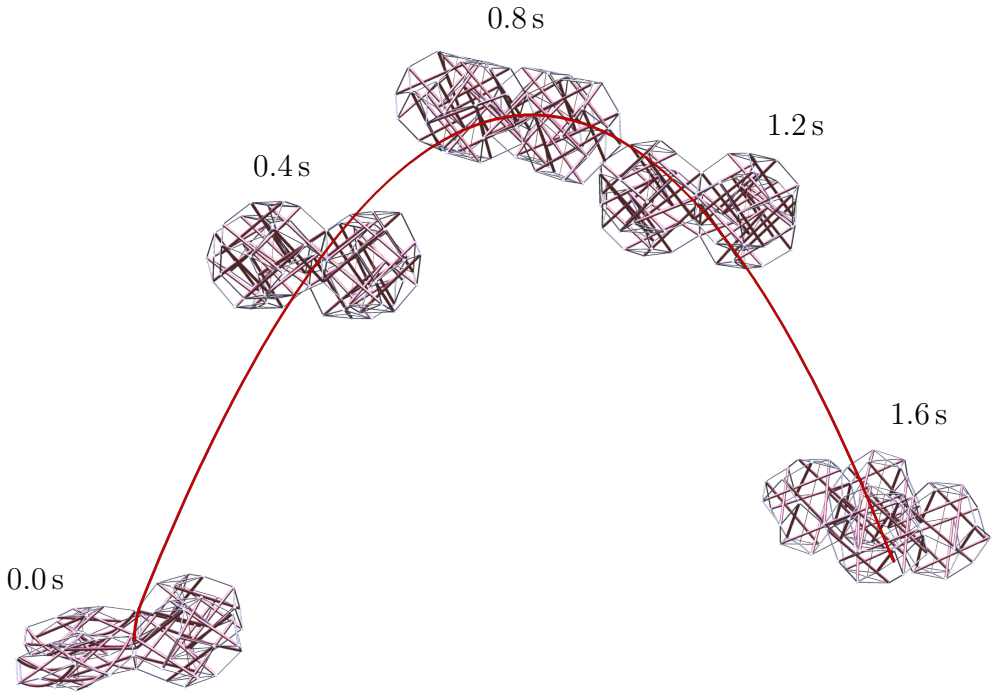

For about \si{10000} equilibrium configurations, hop trajectories were subsequently simulated for \si{3} seconds by unlocking the actuation mechanism. A typical hop trajectory is depicted in Fig.~\ref{fig:jump}. Upon release of the locking mechanism, the device jumps. The height of the jump is related to the combined stretch $\lambda$ of all actuators, whereas the jumping direction depends on the differential stretch between units of the lattice.

\begin{figure}[!ht]
	\centering
	\includegraphics[width=0.6\textwidth]{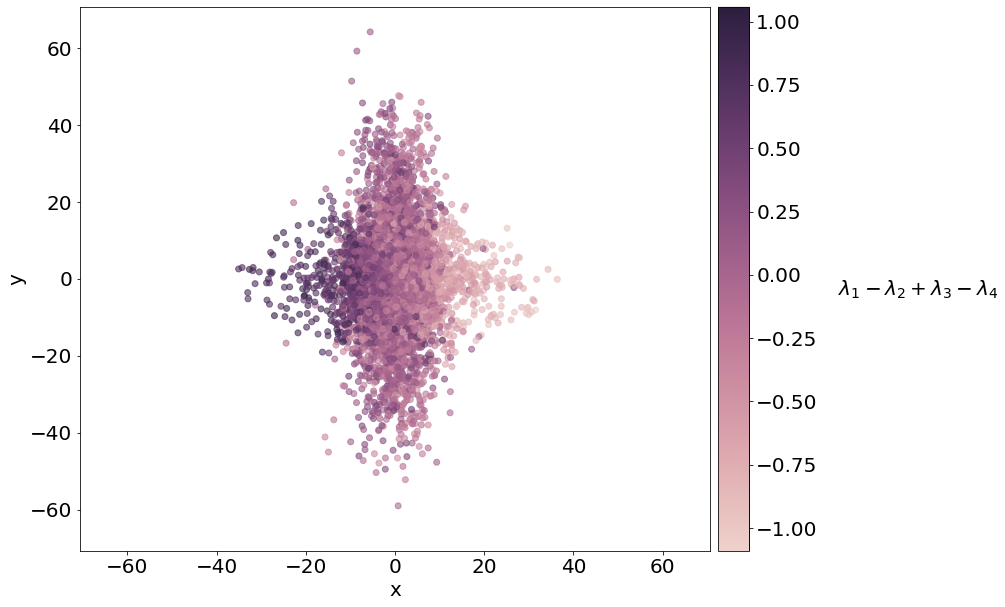}
	\caption{System position after sumlated hopping for about \si{10000} initial actuator configuration. The values of $(\lambda_1, \lambda_2, \lambda_3, \lambda_4)$ used in the color scheme represent the initial retraction of the actuated cables.}
	\label{fig:final_pos:sum_no_relock}
\end{figure}

Results of our simulations are summarized in Fig.~\ref{fig:final_pos:sum_no_relock}. The figure depicts the final position of the center of gravity of the vehicle by the end of the simulation. The position of the center of gravity is approximated by averaging the position of all nodes in the lattice. As it can be seen, the lattice has a better ability to translate in the 0, 90, 180 and 270 degree directions. However, it can still move at all intermediate directions. The figure also highlights the dependence of jump direction on stretch differential, represented by $\lambda_1-\lambda_2+\lambda_3-\lambda_4$. This definition has a four-fold symmetry, so identical results are expected by permuting the stretches while preserving the 90 degree symmetry of the lattice.

\begin{figure}[!ht]
	\centering
	\includegraphics[width=0.6\textwidth]{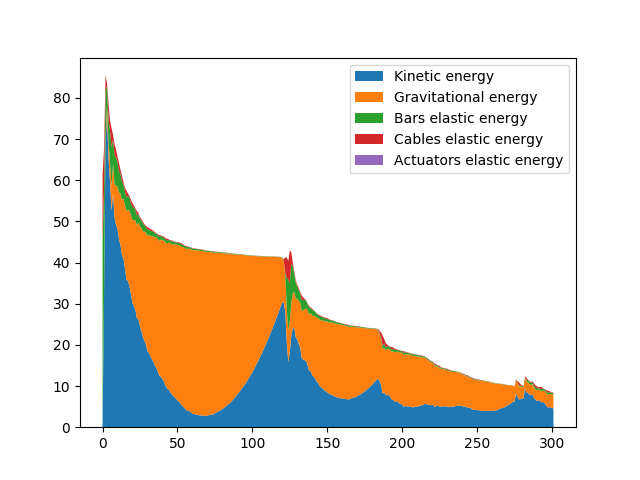}
	\caption{Plot of the energies of the tensegrity system during a simulated hop. The elastic energy stored in the actuating cables, although mentionned in the legend, is too small to be distinguishable on the plot.}
	\label{fig:energies}
\end{figure}

To conclude this section, from the analysis of the simulated trajectories, one can observe that a significant proportion of the elastic energy stored in the tensegrity structure is dissipated because of the friction with the ground. The rest of this energy that is converted in kinetic energy could potentially be converted back to elastic energy at impact with the appropriate relocking mechanism of the actuators. Fig.~\ref{fig:energies} shows the energy of the tensegrity system over time.

%% file: sections/experiments.tex
To illustrate a feasible actuation mechanism able to produce a jump of the tensegrity structure, a $2\times1$ lattice structure was built. The structure consists of a total of 44 nodes, 24 bars, and 84 cables. Each node is covered in reflective tape to be tracked in real time by 12 Vicon cameras. Tracking of the individual nodes provides not only valuable information about of the dynamics the lattice during a flip, but also a better understanding of compression and tension on its bars and cables.

\begin{figure}[!ht]
    \centering
    \begin{subfigure}{0.49\textwidth}
        \centering
        \includegraphics[width=\linewidth]{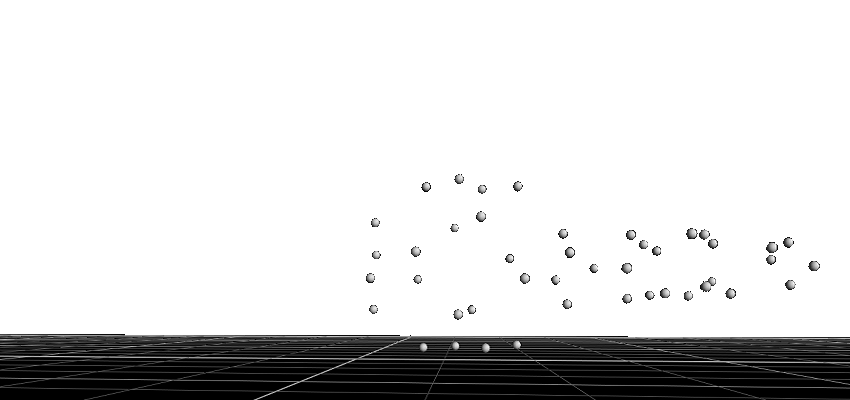}
    \end{subfigure}
    \begin{subfigure}{0.49\textwidth}
        \centering
        \includegraphics[width=\linewidth]{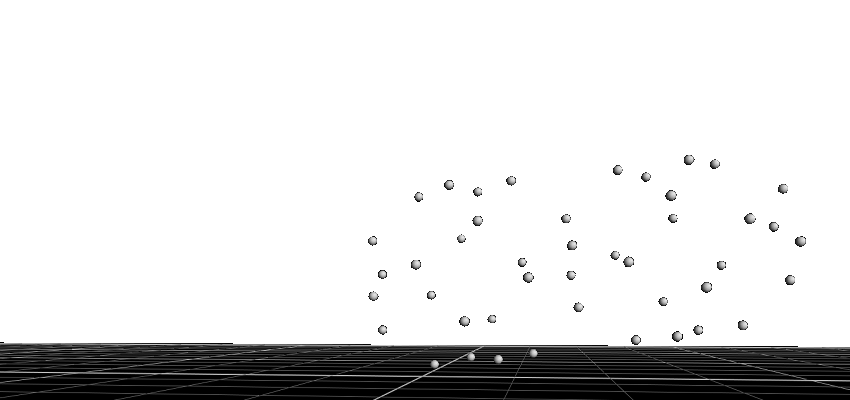}
    \end{subfigure}
    \begin{subfigure}{0.49\textwidth}
        \centering
        \includegraphics[width=\linewidth]{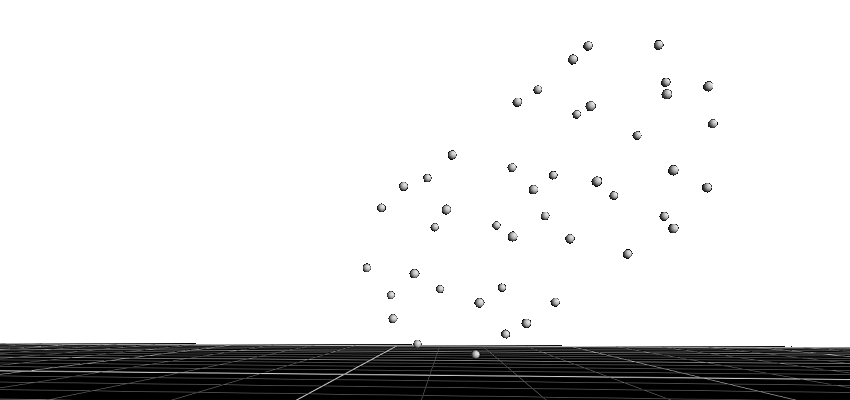}
    \end{subfigure}
    \begin{subfigure}{0.49\textwidth}
        \centering
        \includegraphics[width=\linewidth]{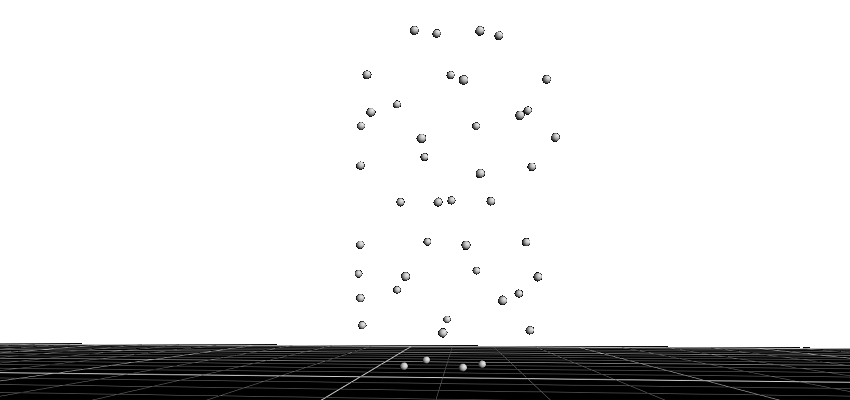}
    \end{subfigure}
    \begin{subfigure}{0.49\textwidth}
        \centering
        \includegraphics[width=\linewidth]{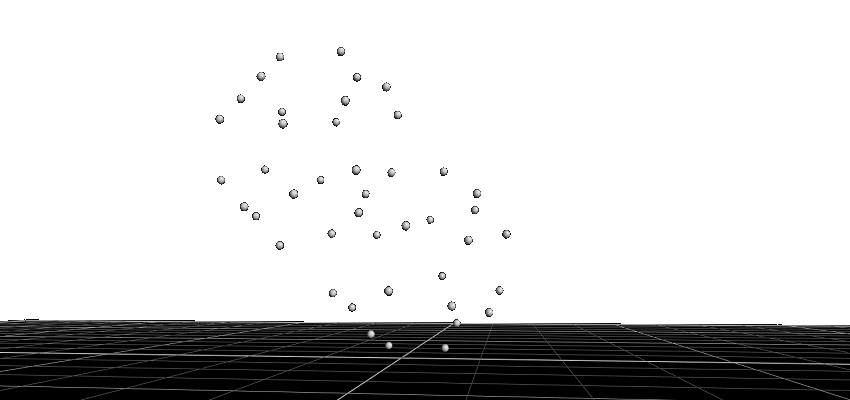}
    \end{subfigure}
    \begin{subfigure}{0.49\textwidth}
        \centering
        \includegraphics[width=\linewidth]{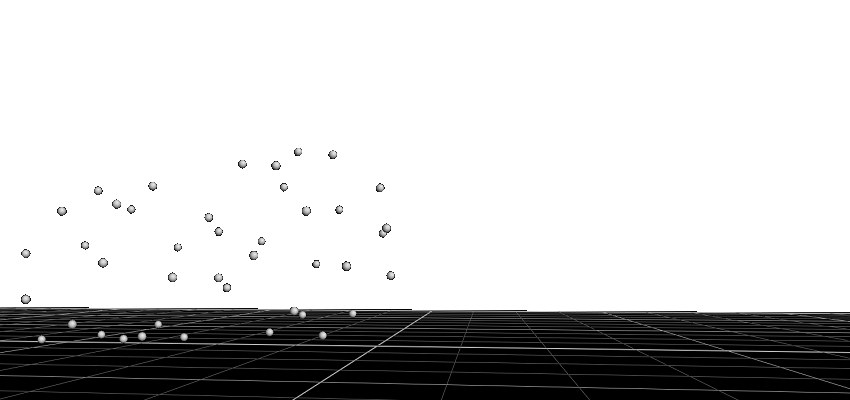}
    \end{subfigure}
    \begin{subfigure}{0.49\textwidth}
        \centering
        \includegraphics[width=\linewidth]{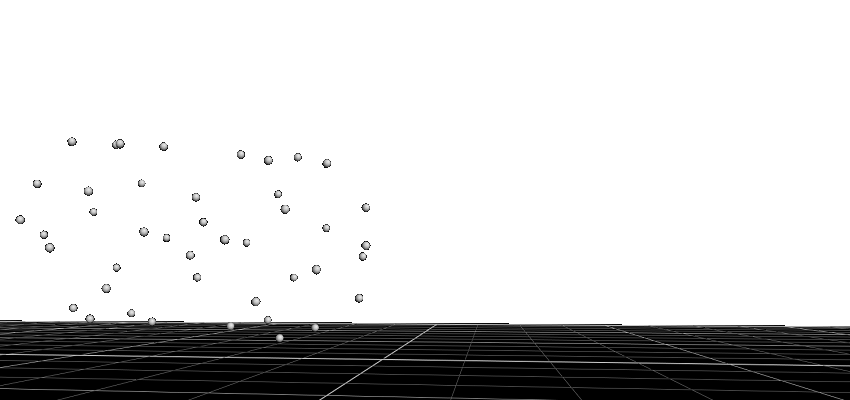}
    \end{subfigure}
    \begin{subfigure}{0.49\textwidth}
        \centering
        \includegraphics[width=\linewidth]{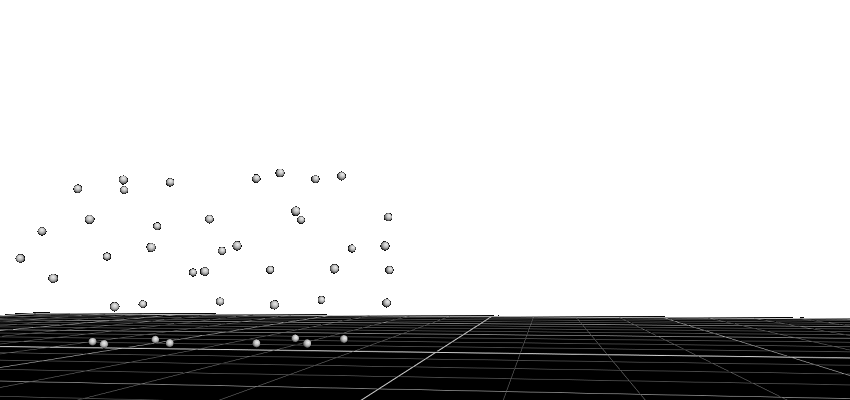}
    \end{subfigure}
    \caption{Snapshots of the positions of the nodes tracked by the Vicon system during the physical experiment at different times. Chronological order is from left to right then top to bottom.}
    \label{fig:vicon}
\end{figure}

The experimental setup using the 2-unit lattice comprised of placing the lattice within a square area that was calibrated to ensure accurate positioning from all cameras. The lattice was located on a variable incline, that was elevated to different positions during experimentation. Before the test began, one unit of the lattice was compressed to the desired state using the actuator mechanism. When fully compressed, an Arduino was powered on to control a servo able of removing a locking mechanism within the actuator. With the lock now removed, the unit was able to release its stored energy, initiating a flip of the 2-unit lattice. Results from one experiment are documented in Fig.~\ref{fig:vicon}, which shows screenshots taken from the Vicon Nexus software used to track the nodes.

Trials were conducted at three different inclinations of 10, 15, and 20 degrees, to evaluate the response of the lattice under varying operational conditions. In this context, the inclination was the angle between the ground and the surface of the ramp the lattice structure was initially placed on. Compression of the lattice was also varied at each inclination, where the actuating cable was retracted to 70\%, 85\%, and 100\% of its initial length of 0.35 meters. These compression values correspond to values of stretch $\lambda$ of approximately 0.52, 0.40, and 0.32 respectively, where $\lambda$ is defined as the compressed lattice height divided by the original, uncompressed height. Three experiments were completed at each combination of inclination and compression, leading to a total of 27 trials.

\begin{figure}[!ht]
	\centering
	\begin{subfigure}{0.49\textwidth}
		\centering
		\includegraphics[width=\linewidth]{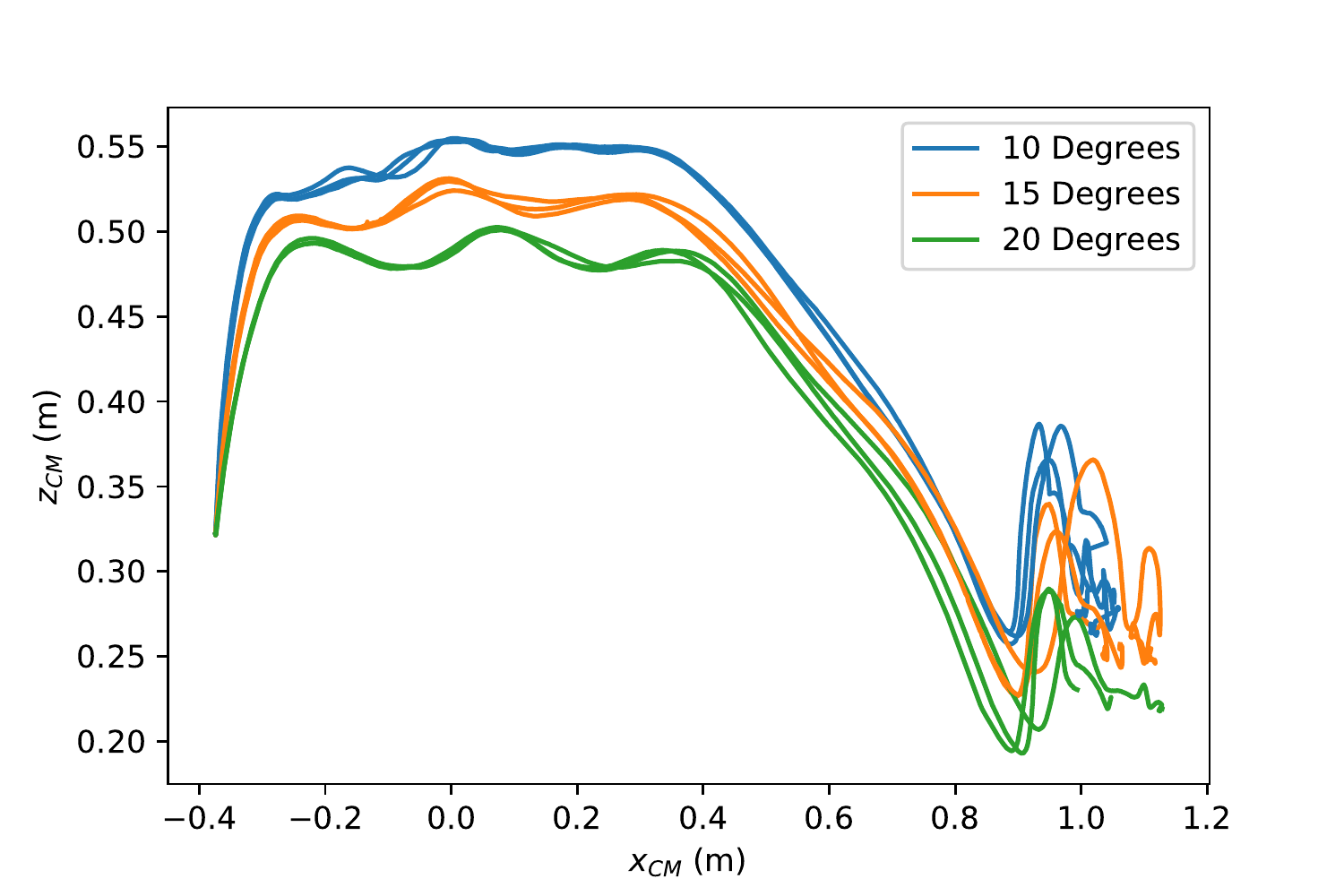}\subcaption{$\lambda=0.32$}
	\end{subfigure}
	\begin{subfigure}{0.49\textwidth}
		\centering
		\includegraphics[width=\linewidth]{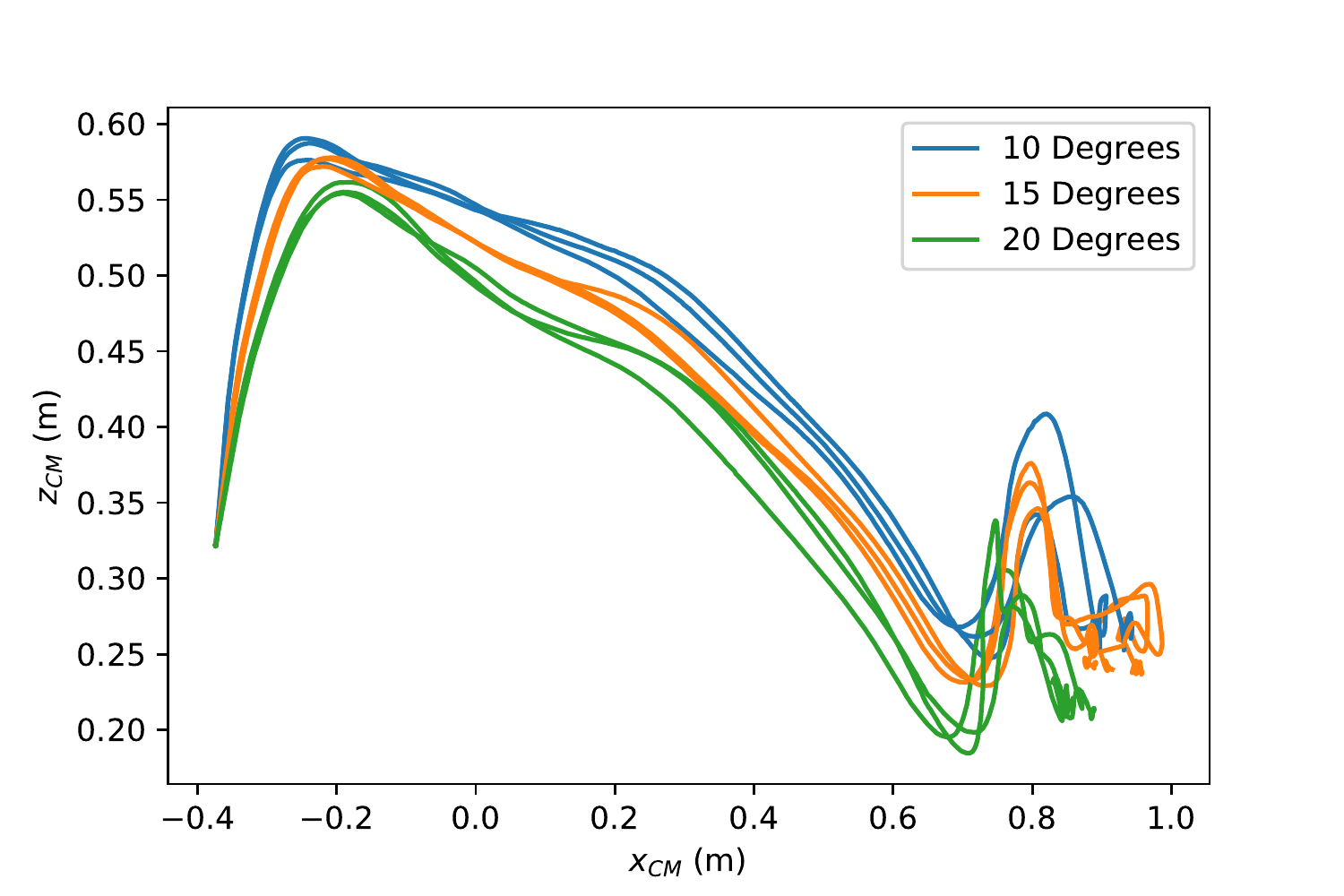}\subcaption{$\lambda=0.40$}
	\end{subfigure}
	\begin{subfigure}{0.49\textwidth}
		\centering
		\includegraphics[width=\linewidth]{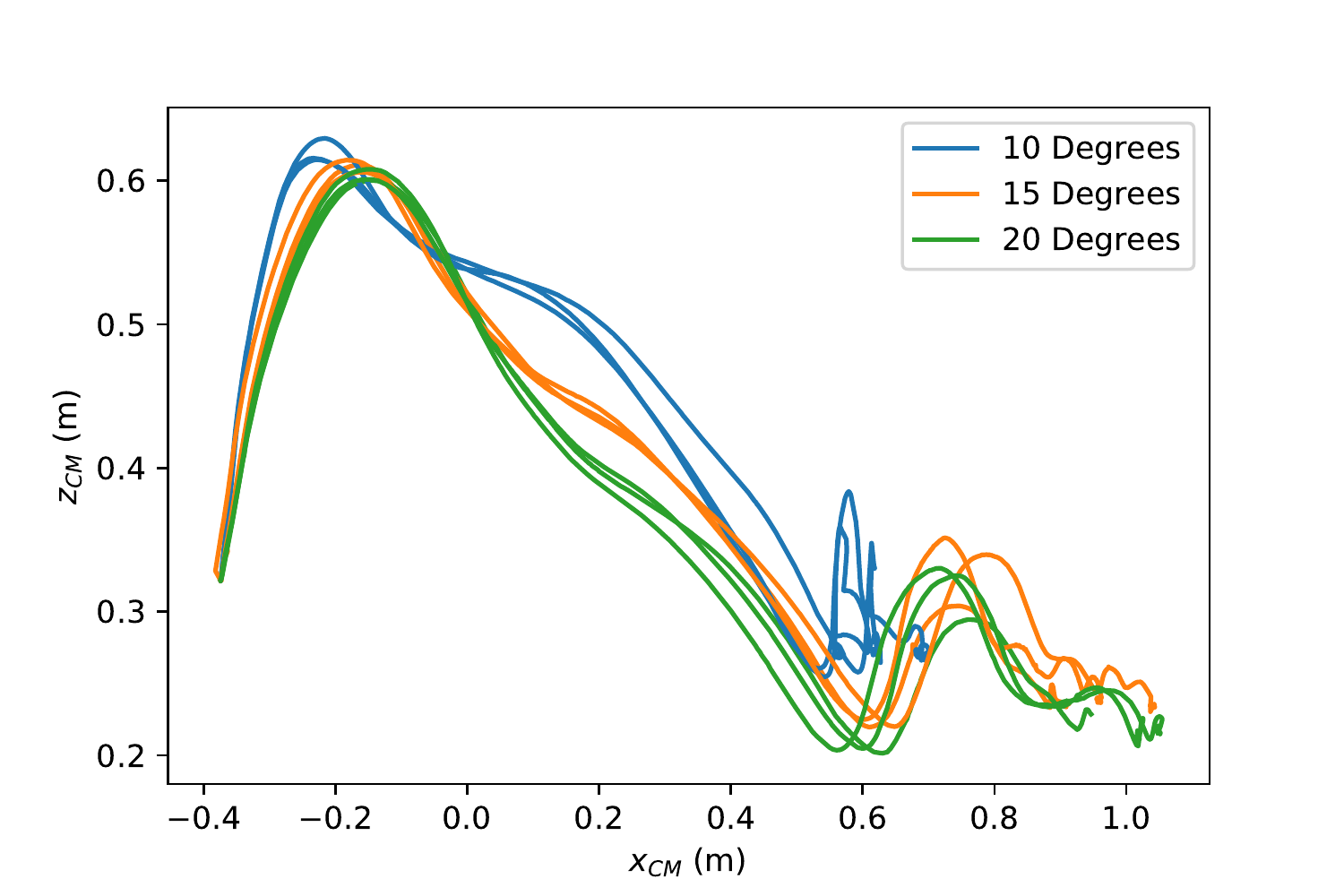}\subcaption{$\lambda=0.52$}
	\end{subfigure}
	\caption{Experimentally obtained position of center of mass for various cable retraction levels.}\label{fig:vicon_cg_history}
\end{figure}
\newpage

\begin{figure}
	\centering
	\includegraphics[width=0.7\textwidth]{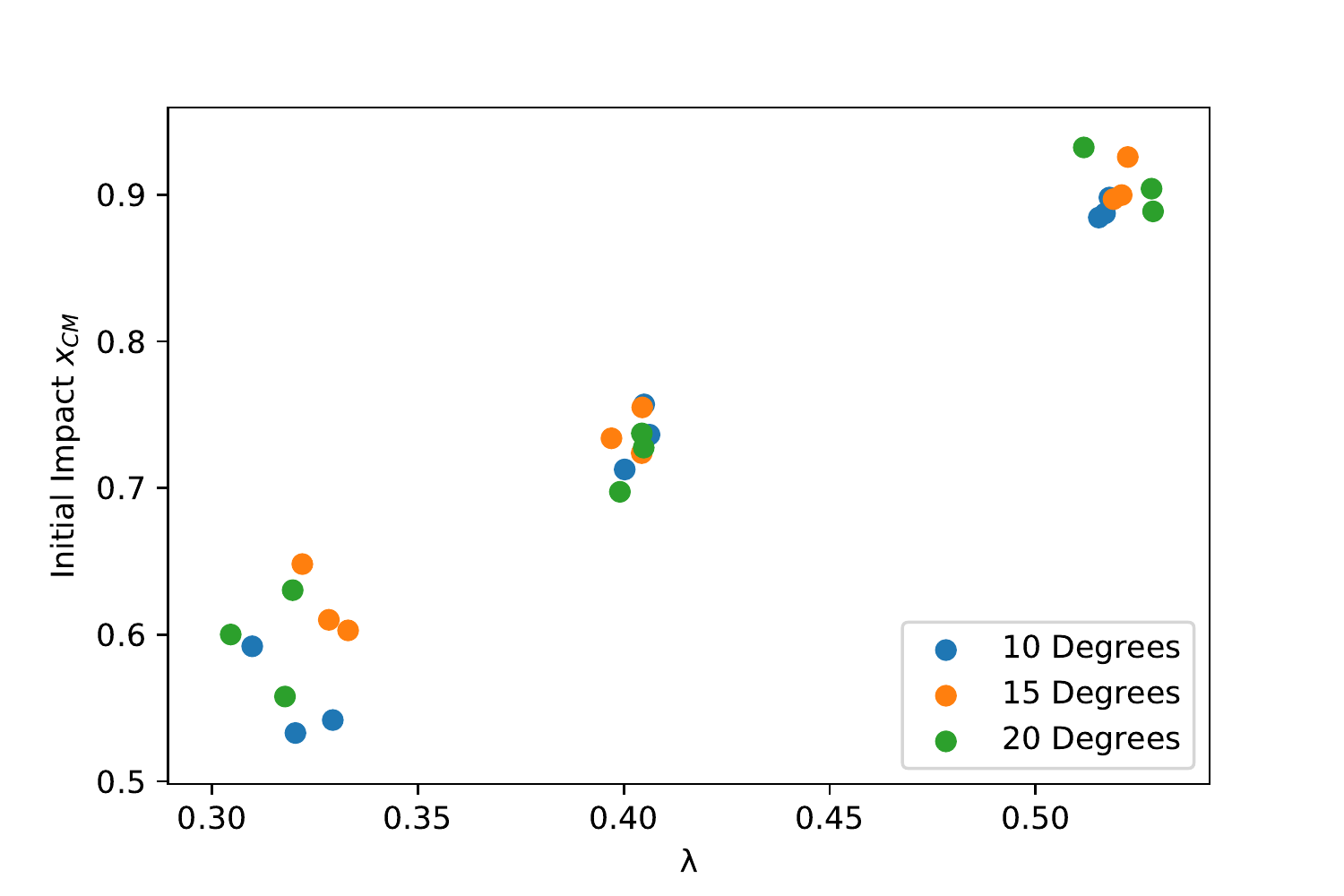}
	\caption{Experimentally obtained position of center of mass at the time of first impact for all experiments.}\label{fig:vicon_cg_impact}
\end{figure}

Fig.~\ref{fig:vicon_cg_history} shows the experimentally obtained position of center of mass for various cable retraction levels. As it can be seen, the incline angle does impact the overall trajectory without affecting significantly the landing location of the center of mass after the flip maneuver. This weak correlation with incline angle can be seen more clearly in Fig.~\ref{fig:vicon_cg_impact}, which shows the position of the structure center of mass at the time of first impact for varying cable retraction levels and incline angles.
Perhaps more interestingly, the figure indicates that horizontal distance traveled by the structure increases as compression is decreased. This result could be somehow counter-intuitive, as one would assume that higher retraction levels are associate with larger energy storage, resulting in longer travel distances. We believe this effect could be due to higher compressed cases becoming airborne for a longer period of time, a state in which the lattice rotates about its center of mass, losing horizontal velocity compared to the lower compressed cases. For the lower compressed trials, one unit remains in contact with the ground for the duration of the test, allowing the lattice to travel further horizontally by exploiting friction with the ground. This leads to an initial conclusion that in order to acquire a maximum horizontal distance traveled, the energy to provide a flip of the lattice should be optimized accounting for friction conditions.

%% file: sections/conclusion.tex
In this work, a new mode of locomotion for tensegrity systems is proposed and analyzed via a reduced-order model suitable to the modeling of the buckling behavior of compression elements. 
This mode of locomotion, based on a hopping motion of the system, is appropriate for tensegrity lattices constructed from a truncated tetrahedron unit cell. 
The actuation mechanism is designed to compress each unit cell of the lattice and quickly release the stored elastic energy to initiate a jump. 
The hopping behavior of a $2\times 2$ lattice is studied by simulating the structure with the presented reduced-order model.
The simulation of the hopping behavior of the tensegrity system is done in two steps involving a conjugate gradient descent to find the equilibrium of a structure given its actuators configuration and a simulation of the structure's dynamics once the the elastic energy stored with the actuation mechanism is released.
A $2\times 1$ lattice prototype was built and subject to various jump tests to validate this novel mode of locomotion. Results show that compression level highly conditions travel distance during the flip maneuver, and that appropriate compression level should be optimized for ground friction conditions. This should be addressed in future work, in which we will aim at refining the simulation parameters to replicate the behavior of a $2\times 2$ lattice. The dataset of trajectories will also be increased in size and learning methods will be studied to predict the trajectory behavior for a given configuration of the actuated cables without using the simulation.